\documentclass[11pt,a4paper]{article}
\usepackage{times,latexsym}
\usepackage{url}
\usepackage[T1]{fontenc}

\usepackage[acceptedWithA]{tacl2018v2}

\usepackage{xspace,mfirstuc,tabulary}

\usepackage{microtype}
\usepackage{graphicx}
\usepackage{amsmath}
\usepackage{amssymb}
\usepackage{booktabs}
\usepackage{bbm}
\usepackage{paralist} % also compact lists
\usepackage{comment}
\usepackage[ruled,vlined]{algorithm2e}
\usepackage{todonotes}
\usepackage{xcolor}

\newcommand{\bi}{\begin{list}{$\bullet$}{
    \setlength{\leftmargin}{1.5 em}
    \setlength{\itemsep}{0 pt}
    \setlength{\topsep}{3 pt}
    \setlength{\parsep}{3 pt}
    \setlength{\partopsep}{0 pt}
    \setlength{\labelwidth}{1 em}
    \setlength{\labelsep}{0.5 em}
    \setlength{\parskip}{0cm}  }}
\newcommand{\ei}{\end{list}}

\newcommand{\BE}{\begin{enumerate}}
\newcommand{\EE}{\end{enumerate}}

\newcommand{\initab}{                           % set up tab stops
\begin{tabbing}
XXX \= XXXX \= \kill
}
\newcommand{\begpub}{
\begin{quotation}
\noindent
}

\newcommand{\finpub}{
\end{quotation}
}

\newcommand\sys{m$\mathcal{B}$-{\sc MAPO}}

\newcommand{\newtext}[1]{\textcolor{red}{#1}}

\newif\iftaclinstructions
\taclinstructionsfalse % AUTHORS: do NOT set this to true
\iftaclinstructions
\renewcommand{\confidential}{}
\renewcommand{\anonsubtext}{(No author info supplied here, for consistency with TACL-submission anonymization requirements)}
\newcommand{\instr}
\fi

\iftaclpubformat % this "if" is set by the choice of options

\else

\fi

\title{Unsupervised Learning of KB Queries in Task-Oriented Dialogs}

\author{
 Dinesh Raghu\thanks{\ \ D. Raghu is an employee at IBM Research. This work was carried out as part of PhD research at IIT Delhi.} $^{\, 1\, 2}$, 
Nikhil Gupta\thanks{\ \ This work was done while Nikhil Gupta was a graduate student at IIT Delhi.} $^{3}$ and
Mausam$^1$ \\
$^1$ IIT Delhi, New Delhi, India\\
$^2$ IBM Research, New Delhi, India\\
$^3$ LimeChat, Gurgaon, India\\
{\em diraghu1@in.ibm.com,
nikhil@limechat.ai,
mausam@cse.iitd.ac.in}
}

\date{}

\begin{document}
\maketitle
\begin{abstract}
Task-oriented dialog (TOD) systems often need to formulate knowledge base (KB) queries corresponding to the user intent and use the query results to generate system responses. Existing approaches require dialog datasets to explicitly annotate these KB queries -- these annotations can be time consuming, and expensive. In response, we define the novel problems of predicting the KB query and training the dialog agent,  without explicit KB query annotation. For query prediction, we propose a reinforcement learning (RL) baseline, which rewards the generation of those queries whose KB results cover the entities mentioned in subsequent dialog. Further analysis reveals that correlation among query attributes in KB can significantly confuse memory augmented policy optimization (MAPO), an existing state of the art RL agent. To address this, we improve the MAPO baseline with simple but important modifications suited to our task.
%To address the issue, we propose simple modifications to an existing policy optimization technique.
%To address the issue, we propose simple modifications to an existing policy optimization technique, which in addition to on-policy samples, uses a buffer of highest reward queries seen during training to estimate the gradients.

To train the full TOD system for our setting, we propose a pipelined approach: it independently predicts when to make a KB query (query position predictor), then predicts a KB query at the predicted position (query predictor), and uses the results of predicted query in subsequent dialog (next response predictor). 
%We then propose an approach to learn TOD system using unannotated dialogs by utilizing the RL-based query predictor. 
Overall, our work proposes first solutions to our novel problem, and our analysis highlights the research challenges in training TOD systems without query annotation.

%in bridging the performance gap between the existing state-of-the-art approaches that learn TOD systems using query-annotated dialogs and our baseline, which learns TOD systems without query annotation.}

%Task-oriented dialog (TOD) systems are designed to assist users in accomplishing specific tasks. These tasks often require the system to formulate a knowledge base (KB) query corresponding to the user intent and use the query results to generate system responses. In order to train end-to-end TOD systems, dialogs are required to be manually annotated with these KB queries. Such annotations are time consuming, expensive and hinder scalabilty. To alleviate the need for such annotations, we define the problem of learning KB queries in an unsupervised manner using reinforcement learning (RL). Using existing policy optimization techniques to train the RL policy is uniquely plagued by correlated attributes problem, in which, due to strong correlations between query attributes in TOD, RL agent fails to learn a good policy. To address this, we propose correlated attributes resilient gradient estimation (\sys), which in addition to on-policy samples, also uses a buffer of highest reward samples seen during training to estimate the gradients. We show that \sys\ achieves better KB query prediction accuracy compared to existing gradient estimation techniques on two standard TOD datasets. We also compare and analyze performance of end-to-end TOD systems trained using dialogs with KB queries annotated by various RL approaches.
\end{abstract}

\section{Introduction}

\begin{figure*}[t]
\centering
\includegraphics[width=0.9\textwidth]{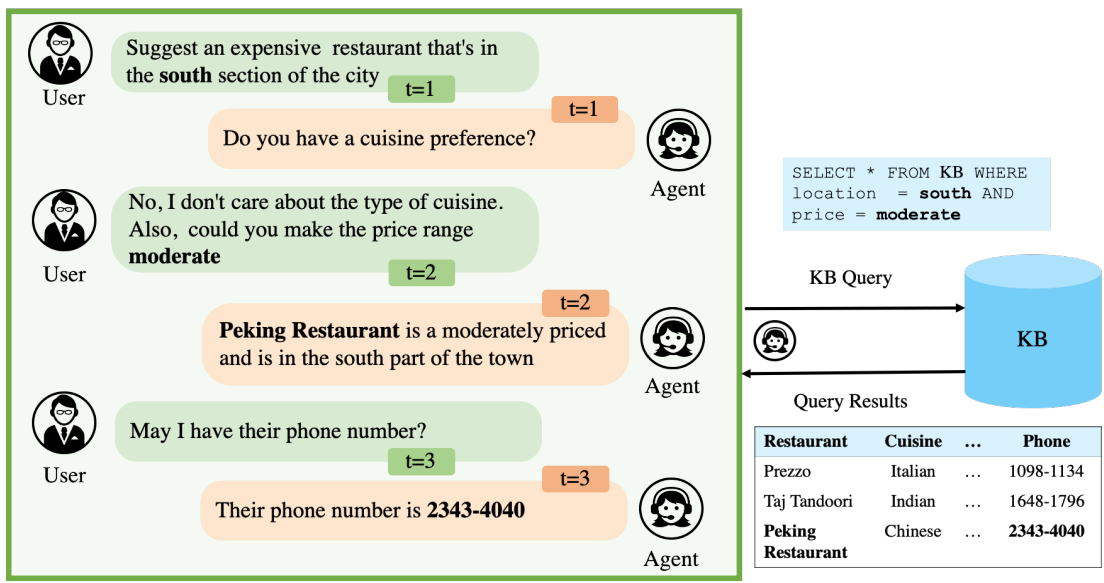}
\caption{ Example of a training task-oriented dialog. At turn 2, the agent first queries the KB based on the user requirement and then responds based on the retrieved results.}
\label{fig:examples}
\end{figure*}

Task-oriented dialog (TOD) systems converse with users to accomplish specific tasks such as restaurant reservation \cite{hen2014word}, movie ticket booking \cite{li2017end} or bus enquiry \cite{Raux2005LetsGP}. In addition to the ability to converse, it is crucial for TOD systems to learn to formulate knowledge base (KB) queries based on the user needs, and generate responses using the query results. An example task oriented dialog is shown in Figure \ref{fig:examples}, where during the conversation (at turn 2), the agent queries the KB based on the user needs, and then suggests the \textit{Peking Restaurant} based on the retrieved results. Existing end-to-end approaches~\cite{BordesW16,mem2seq,reddy2019multi} learn to formulate KB queries using manually annotated queries.

In real-world scenarios, human agents chat with users on messaging platforms. When the need to query the KB arises, the agent fires the query on a back-end KB application and uses the retrieved results to compose a response back on the messaging platform.  The dialogs retrieved from these platforms contain just the user and the agent utterances, but the KB queries typically go undocumented. As existing approaches require KB annotations, they have to be manually annotated, which is expensive and hinders scalability. To eliminate the need for such annotations, we define the novel problem of training a TOD system without explicit KB query annotation. A key subtask of such a system is the unsupervised prediction of KB queries.

%predicting KB queries in an unsupervised manner without explicit KB annotation and use the predicted queries to train a TOD system.
%The predicted queries can then be used to train an end-to-end TOD system. 

While there is no explicit query annotation, we observe that dialog data still offers \emph{weak supervision} to induce KB queries -- all the entities used by the agent in subsequent dialog should be returned by the correct query. For example, in Figure \ref{fig:examples}, the correct query should retrieve \textit{Peking restaurant} and its phone number. This suggests a reinforcement learning (RL) formulation for query prediction, not unlike similar tasks in Question Answering (QA) \cite{artzi2013weakly} and conversational QA \cite{yu2019cosql}. However, there is one key difference between TOD and QA settings -- in TOD only a few entities from the query results are used in subsequent dialog, whereas in QA all correct answers are provided at training time. This makes defining the reward function more challenging for our setting.

Our problem is further exacerbated by the issue of \textit{correlated attributes}. Query attributes in TOD may exhibit significant correlation, like in the restaurant domain, cuisine and price range are often correlated. For example, most Japanese restaurants in a KB may be in expensive price range. As a result, presence and absence of expensive price range in the query could retrieve almost the same set of KB entities and hence similar rewards. This can confuse the weakly supervised query predictor. 
%Second, \textit{variation in dialogs for similar intents} -- two dialogs with the same user intent can have different sets (and sizes) of KB entities mentioned in the subsequent dialogs. It is important to define a reward function that normalizes across such settings, or else one of the two dialogs might dominate at learning time. 
To counter this issue we present a baseline solution for KB query prediction by extending an existing policy optimization technique, MAPO \cite{liang2018memory}. Experiments show that our proposed modification significantly improves the query prediction accuracy.

To train a full TOD system without KB query annotation, we propose a pipelined solution. It uses three main components: (1) query position predictor predicts when a query must be made, (2) query predictor predicts the query at the turn predicted by position predictor, and (3) next response predictor generates next utterance based on dialog context, and predicted query's results. We train these components in a curriculum, due to the pipeline nature of the system. We find that overall our system obtains good dialog performance, and also learns to generalize to entities unseen during train.

We conclude with novel research challenges highlighted by our paper. In particular, we notice that even fully supervised TOD systems (trained with KB query annotation) suffer a significant loss in performance, when they are evaluated with the protocol of using their own predicted queries and its query results (instead of using gold queries and gold results) in subsequent dialog. We believe that designing TOD systems with high performance under this evaluation protocol is the key next step towards making end-to-end TOD systems useful in real applications. We release all our resources for further research -- training data, evaluation code, and code for TOD systems\footnote{\url{https://github.com/dair-iitd/mb-mapo}}.

%\newtext{Predicting the correct KB query is crucial for learning a TOD system. The system generates response based on the dialog so far and the query results. Without the correct KB query, the results would not contain the required KB entities and hence the system would memorize the KB entities in the response, rather than inferring them from the query results. Such a system would not generalizes to support entities unseen during train. We propose an approach to learn TOD system using dialogs without annotations by utilizing the query predictor. We show that our proposed approach learns to generalize to entities unseen during train. We finally discuss open challenges involved in bridging the gap between the existing approaches for learning TOD systems using dialogs with annotations and the proposed approach for learning them using dialogs without annotations.}

%To summarize, we make the following contributions: 
%\begin{compactenum}
%    \item We define the problem of learning KB query annotations in an unsupervised manner using RL.
%    \item We propose a novel correlated attributes resilient gradient estimation (\sys) technique for predicting KB queries.
%    \item We show that our proposed approach, \sys, outperforms existing approaches on KB query prediction in TOD.
%\end{compactenum}

\section{Background \& Related Work}

Our work is on task-oriented dialogs and is closely related to the task of semantic parsing. We briefly discuss related work in both areas. Our query predictor is an extension of memory augmented policy optimization (MAPO) \cite{liang2018memory}, which we describe in detail.

\vspace{0.5ex}
\noindent \textbf{Task Oriented Dialogs}: TOD systems are of two main types: traditional spoken dialog systems (SDS) and end-to-end TOD systems. SDS \cite{wen2017network,williams2017hybrid} use hand crafted states and state annotations on every utterance in the dialogs -- a significant human supervision. End-to-end TOD systems \cite{reddy2019multi,wu2019global,raghu2019disentangling} do not require state annotations but just the KB query annotations. There exist approaches \cite{chen2013unsupervised,chen2015matrix} to induce state annotations  in SDS, but we are the first to induce query annotations in end-to-end TOD systems. TOD systems cannot be learnt with just the state annotations, additional state to KB query mapping/annotation is required. But no further annotations are needed to learn TOD system when query annotations are available.

We build on recent architectures, which use memory networks to store previous utterances and query results, and a generative copy decoder to construct the agent utterances \cite{mem2seq,reddy2019multi,raghu2019disentangling}. An alternative approach maintains the entire KB in its neural model, bypassing the need for KB queries entirely \cite{dhingra2017towards,Ericsigdial}. Unfortunately, such approaches can only work with small KBs. In contrast, our approach is scalable and does not impose restrictions on KB size.

\vspace{0.5ex}
\noindent \textbf{Semantic Parsing}: \textcolor{black}{Semantic parsing is the task of mapping natural language text to a logical form (or program) \cite{zelle96,zettlemoyer05}. To alleviate the need for gold program annotations, weakly supervised approaches have been proposed \cite{artzi2013weakly,berant13,pasupat-liang-2015-compositional,DBLP:conf/iclr/NeelakantanLAMA17,haug2018neural}. These weakly supervised approaches solve two main problems: (1) exploring large search space to find correct logical programs and (2) spurious program problem -- where the policy learns an incorrect program that fetches the correct answer. To explore large search space better, \newcite{guu2017language} used randomized beam search, \newcite{liang2017neural} and MAPO used systematic search. There exists several  approaches to overcome the spurious query problem: crowd sourcing \cite{pasupat-liang-2016-inferring}, spreading the probability mass over multiple reward earning programs \cite{guu2017language}, clustering similar natural language inputs and using their abstract representations \cite{goldman-etal-2018-weakly},and using overlap between the text spans in input and the generated programs \cite{wang2019learning,dasigi-etal-2019-iterative,misra2018policy}. MAPO samples from a buffer of systematically explored reward earning programs based on their likelihood in the current policy \cite{liang2018memory} to tackle the spurious program problem. Our query prediction approach follows this literature, except in a dialog setting. As our approach is an extension of MAPO, we inherit the ability to effectively tackle the two issues.}

These methods use an RL formalism, in which a logical form (query, in our case) $\textbf{a}$ is predicted by an RL-policy $\pi_\theta(\textbf{a}|c)$. Here, $\theta$ are the parameters of the RL agent, and $c$ is the input, e.g., a question (in our case, a dialog context).
The policy $\pi_\theta$ is trained by maximizing the expected reward:
\begin{align}
\mathcal{O}_{ER} &= \mathbb{E}_{\textbf{a} \sim \pi_{\theta}(\textbf{a}|c)} R(\textbf{a}|c,y) = \mathbb{E}_{\textbf{a} \sim \pi_{\theta}(\textbf{a})} R(\textbf{a}) \notag
%&= \sum_{\textbf{a} \in \mathcal{A}} \pi_{\theta}(\textbf{a}|c) R(\textbf{a}|c)
\end{align}
%where $\mathcal{A}$ is the set of all possible queries, and $R$ is the reward function.  
where $R$ is the reward function and $y$ is the gold answer for $c$. For simplicity, we drop the dependence of $\pi_{\theta}$ and $R$ on $c$ and $y$. REINFORCE \cite{williams1992simple} can be used to estimate the gradient of the expected reward. Using $N$ queries sampled i.i.d. from the current policy, the gradient estimate can be expressed as:
\begin{align}
\nabla_{\theta} \mathcal{O}_{ER} = \frac{1}{N} \sum_{k=1}^{N} \nabla_{\theta} \text{log} \pi_{\theta}(\textbf{a}_k)  R(\textbf{a}_k) 
\label{eq:reinforce}
\end{align} 
When the search space is large and the rewards are sparse, relying on just the on-policy samples often leads to poor search space exploration. To overcome this, search is added on top of policy samples \cite{liang2017neural}. We build on MAPO, which uses \textit{systematic exploration} to identify non-zero reward queries  for each training data and stores them in a memory buffer $\mathcal{B}$. The expected reward is computed as a weighted sum of two expectations: one over the queries inside the buffer $\mathcal{B}$ and the other over the remaining queries:
\begin{align}
\mathcal{O}_{ER} = \sum_{\textbf{a} \in \mathcal{B}} \pi_{\theta}(\textbf{a}) R(\textbf{a}) + \!\!\!\!\sum_{\textbf{a} \in \mathcal{A} - \mathcal{B}} \pi_{\theta}(\textbf{a}) R(\textbf{a}) \notag \\
= \pi_{\mathcal{B}} \mathbb{E}_{\textbf{a} \sim \pi_{\theta}^{+}(\textbf{a})} R(\textbf{a}) + (1-\pi_{\mathcal{B}}) \mathbb{E}_{\textbf{a} \sim \pi_{\theta}^{-}(\textbf{a})} R(\textbf{a})
\label{eq:mapo}
\end{align}
where $\mathcal{A}$ is the set of all possible queries, $\pi_{\mathcal{B}}=\sum_{\textbf{a} \in \mathcal{B}} \pi_{\theta}(\textbf{a})$ is total probability of all queries in the buffer, and $\pi_{\theta}^{+}(\textbf{a})$ and $\pi_{\theta}^{-}(\textbf{a})$ are the normalized probability distributions inside and outside the buffer respectively.
The gradient of the first term is estimated exactly by enumerating all queries in the buffer, while the gradients of the second term is estimated as in Equation \ref{eq:reinforce} by sampling i.i.d queries from the current policy and rejecting them if they are present in the buffer $\mathcal{B}$. 

A randomly initialized policy is likely to assign small probabilities to the queries in the buffer, and hence negligible contribution to the gradient estimation. To ensure queries in the buffer contribute significantly to the gradient estimates, MAPO clips $\pi_{\mathcal{B}}$ to $\alpha$. The modified gradient estimate is,
\begin{align}
 \nabla_{\theta} \mathcal{O}^{c}_{ER} = \pi^c_{\mathcal{B}} \mathbb{E}_{\textbf{a} \sim \pi_{\theta}^{+}(\textbf{a})} \nabla_{\theta} \text{log} \pi_{\theta}(\textbf{a}) R(\textbf{a}) \notag \\ + (1-\pi^{c}_{\mathcal{B}}) \mathbb{E}_{\textbf{a} \sim \pi_{\theta}^{-}(\textbf{a})} \nabla_{\theta} \text{log} \pi_{\theta}(\textbf{a}) R(\textbf{a})
\end{align}

where $\pi^c_{\mathcal{B}} = \text{max}(\pi_{\mathcal{B}}, \alpha)$. When the training begins, $\alpha > \pi_{\mathcal{B}}$ resulting in gradient estimates biased towards the queries in the buffer. Once the policy stabilizes, $\pi_{\mathcal{B}}$ gets larger than $\alpha$ and the estimates becomes unbiased.

%We propose modifications to MAPO which can helps in better 

%As the space of logical forms is exponentially large and the reward signal is sparse, using  REINFORCE \cite{williams1992simple} to estimate policy gradients fails to effectively explore the search space. To overcome this, \newcite{liang2017neural,liang2018memory} perform a systematic search to identify logical forms yielding high rewards, and use those to estimate  gradients. We build on top of the approach proposed by \newcite{liang2018memory}, described next.
%to further tackle correlated attributes and dialog variations.

\textcolor{black}{
\noindent \textbf{Experience Replay}: Experience replay~\cite{lin-experience-replay} stores past experience in a buffer and reuses that to stabilize training and improve sample efficiency. Prioritized experience replay \cite{Schaul2016} assigns priorities to the experiences and samples them based on the priorities to efficiently learn using them. MAPO and our approach use replay buffers which store non-zero reward queries. These queries include both past experiences and queries identified using systematic search. The sampling strategy is similar to prioritized experience replay, where the priorities are defined by the probability of the query in the current policy.
}

\section{Task Definition \& Baseline System}
We first define our novel task of learning TOD system using \textit{unannotated} dialogs -- we name it uTOD. We then describe the KB query predictor in detail. Finally, we describe our proposed baseline uTOD system, which uses the KB query predictor.

\subsection{Problem Definition}

We represent a dialog $d$
between a user $u$ and an agent $s$ as $\{c_1^u, c_1^s,c_2^u, c_2^s,\ldots,c_m^u, c_m^s\}$ where $m$ denotes the number of turns in the dialog. Our goal is to train a uTOD system, which, for all turns $i$, takes the partial dialog-so-far $\{c^u_1, c^s_1,\ldots,c^u_{i-1}, c^s_{i-1},c^u_{i}\}$ as input and predicts the next system response  $\hat{c}^s_i$. Such a uTOD system is trained by the training data $\mathcal{D}$ comprising complete dialogs $\{d_j\}_{j=1}^{|\mathcal{D}|}$, and an associated knowledge base $KB$.

System responses in a dialog often use certain entities from $KB$. To accomplish this, a uTOD system can fire an appropriate query $\textbf{a}$ to $KB$ and fetch the query results $E^a$, and use those to generate its response. No explicit supervision is provided on the gold queries, at either training or test time. I.e., a uTOD system neither knows which query was fired to $KB$, nor knows when it was fired. \textcolor{black}{In this work we assume that the system can make only one KB query per dialog}. Our overall uTOD baseline uses a pipeline of multiple components --   a query position predictor, a query predictor and a next response predictor.

\subsection{KB Query Predictor}
\label{ssec:query-predictor}
We first assume that we have access to the specific turn number $q, 1\leq q \leq m$, where a KB query gets fired by the system (an assumption we relax in the next section). Thus, formally, we are given the \emph{dialog context}  $c=\{c_1^u, c_1^s,c_2^u, c_2^s,\ldots,c_q^u\}$, and $KB$, and the goal of query predictor is to output $\textbf{a}$ that appropriately captures the user intent. We use the term \emph{subsequent dialog} to refer to all utterances that follow the KB query -- $\{c_q^s, c_{q+1}^u,\ldots,c_m^s\}$. In Figure \ref{fig:examples}, the subsequent dialog starts from the second agent utterance. Let $E^s$ be the set of KB entities present in the subsequent dialog. In Figure \ref{fig:examples}, $E^s = \{\textit{Peking Restaurant}, \textit{2343-4040}\}$. We train the query predictor using RL, by providing feedback based on the predicted query's ability to retrieve the entities in $E^s$.

%We define two segments in a dialog $d$ based on the turn at which the KB query is to be made: (1) All utterances from the start of the dialog till the utterance after which a KB query is made is referred to as the dialog context $c$. In Figure \ref{fig:examples}, $c$ spans till the second user utterance. (2) All utterances that follow the context till either the next turn at which a KB query is to be made or the end of the dialog is referred to as the subsequent dialog. In Figure \ref{fig:examples}, the subsequent dialog starts from the second agent utterance. Let $e_s$ be the set of KB entities present in the subsequent dialog. In Figure \ref{fig:examples}, $e^s = \{\textit{Peking Restaurant}, \textit{2343-4040}\}$. 
%The query predictor induces a KB query based on the context $c$. 

%\subsection{KB Query Predictor}
%In this section, we first define the reinforcement learning formulation. We then describe the two main issues that hinders the RL agent from learning a good policy: \textit{correlated attributes} and \textit{unequal rewards}. Along with each issue, we propose modifications to better estimate the gradients. 
%Finally, we describe the network architecture used for implementing policy defined by Equation \ref{eq:policy}.

%\subsubsection{Reinforcement Learning Formulation}

% describe the state/action

Our proposed baseline follows the literature on semantic parsing with weak supervision using RL, and treats the query prediction as equivalent to learning policy $\pi_\theta(\textbf{a}|c)$ that takes the dialog context $c$ as the input and generates a KB query $\textbf{a} = a_1a_2\cdots a_T$. The KB query is a sequence of $T$ actions, where each action is a word predicted by the query predictor. For a given context, the set of all possible actions is a union of a set of keywords\footnote{\{SELECT, *, =, FROM, AND, WHERE\}} in the SQL query language, set of field names in $KB$, the set of all words in the context $c$ and an \textit{<eoq>} token to indicate the end of query. The query is generated auto-regressively as:
\begin{align}
\pi_\theta(\textbf{a}|c) 
%&= \pi_\theta(a_1, a_2, \ldots, a_T|c) \notag\\
&= \prod_{t=1}^{T}\pi_\theta(a_t|a_{1:t-1}, c)
\label{eq:policy} 
\end{align}
As the environment is deterministic, at each time $t$, the RL state can be fully defined by the dialog context $c$ and the actions predicted so far $a_{1:t-1}$. RL gradients can be computed using REINFORCE or the MAPO algorithm.

The policy network is implemented with a standard encoder-decoder architecture for TOD systems. The context is encoded using a multi-hop memory encoder \cite{sukhbaatar2015end} with a bag of sequences memory \cite{raghu2019disentangling}. In the bag of sequences memory, the context is represented as a set of utterances and each utterance as a sequence of words. Each utterance is encoded using a bi-directional GRU \cite{cho2014learning}.

The KB query is generated one word at a time by a copy-augmented sequence decoder \cite{gu2016incorporating}. The search space is the output space of this variable length, copy-augmented sequence decoder. At each time step, the decoder computes a copy distribution over words in the dialog context and a generate distribution over words KB field names and SQL keywords. Finally, a soft gate~\cite{see2017get} is used to combine the two distributions and a word is sampled from the combined distribution. By allowing only copying words from the dialog context to the values in the SQL WHERE clauses, we reduce the decoder's ability to predict spurious queries. Following ~\newcite{ghazvininejad2016generating}, we use beam search guided by the SQL grammar to generate only syntactically correct queries. 

\subsubsection{Reward Functions}
% describe reward
We now describe a vanilla reward function considered for training the RL agent.  For a given $(c, E^s)$ pair, the query $\textbf{a}$ is predicted using $c$ and the reward is computed using $E^s$. Since partial query isn't meaningful, the reward is computed only at time $t=T$, when the complete query is generated. 
%For ease of notation, 
%we drop dependence of the policy on $c$  and use $\pi_{\theta}(\textbf{a})$ to mean $\pi_{\theta}(\textbf{a}|c)$. We similarly 
%use $R(\textbf{a})$ as a shorthand for $R(\textbf{a}|c)$.
%and $R(\textbf{a})$ respectively.

Let $E^a$ be the set of entities retrieved by the query $\textbf{a}$, then the reward at time $T$ must ensure that  all entities present in the subsequent dialog are retrieved by the query. Moreover, the query should be penalized for retrieving more than required entities.  This can be  operationalized as follows:
\begin{align}
R(\textbf{a}|c, E^s) = {\mathbbm{1}}_{\text{recall}(E^s, E^a)=1} . \text{prec}(E^s, E^a)
\label{eq:reward} 
\end{align}

The recall based indicator  ensures that all entities are retrieved, and precision penalizes retrieval of a large number of entities. While a reasonable first solution, our initial experiments showed that MAPO with this reward function does not achieve satisfactory results. We now discuss the challenging aspects of our task, and our improved baseline, \emph{multi buffer} MAPO, to resolve those issues.

\begin{table*}[ht]
\centering
\small
\begin{tabular}{@{}cllll@{}}
\toprule
\textbf{\begin{tabular}[c]{@{}c@{}} User \\ Intent \end{tabular}} &
\textbf{(a)} &
\begin{tabular}[c]{@{}l@{}}user needs a restaurant that serves \\ Chinese with moderate price range\end{tabular}  & 
\textbf{(b)} &
\begin{tabular}[c]{@{}l@{}}user needs a Japanese restaurant \\ in moderate price range \end{tabular} \\
\midrule
\textbf{\begin{tabular}[c]{@{}c@{}} KB \\ Queries \end{tabular}} & &
\begin{tabular}[c]{@{}l@{}}1. cuisine=\textit{chinese} AND price=\textit{moderate} \\ 2. price=\textit{moderate} AND cuisine=\textit{chinese} \\  3. cuisine=\textit{chinese} \\ 4. price=\textit{moderate} \end{tabular} & &
\begin{tabular}[c]{@{}l@{}}1. cuisine=\textit{japanese} AND price=\textit{moderate} \\ 
2. price=\textit{moderate} AND cuisine=\textit{japanese} \\ 
3. cuisine=\textit{japanese} \\ 4. phone=\textit{98232-66789} \end{tabular}  \\
\bottomrule
\end{tabular}
\caption{Summary of dialogs and a few examples of non-zero reward KB queries for the corresponding dialogs. For simplicity, the SELECT clauses are removed from the KB queries.}
\label{tab:ca}
\end{table*}

\subsubsection{Multi Buffer MAPO (m$\mathcal{B}$-MAPO)}
Our problem can become particularly challenging if $KB$ has correlated attributes. To understand this, consider two types of KB queries:  (1) partial intent query and (2) complete intent query. A partial query is one that is partially correct, i.e., captures a part of the user's intent, and does not include any incorrect clauses. A complete query contains all (and only) correct clauses expressed in the intent. For example (a) in Table \ref{tab:ca}, the first two queries are complete ones, whereas rest are partial. Query 4 in column (b) is a spurious query that can retrieve the same result as the complete intent query. This spurious query is not a partial query, as it does not capture any of the user needs. Our query decoder allows only words to be copied from the context when predicting the query, thus reducing the policy network's ability to learn spurious queries.

%Because of the nature of weak supervision in uTOD, it is possible that the entities mentioned in subsequent dialog be covered by multiple queries.
Because of the nature of weak supervision in uTOD (we get to see only a subset of entities in subsequent dialog), it is possible that multiple queries can achieve non-zero rewards for a dialog context. It is further possible that a \emph{specific} partial query can achieve non-zero rewards in many dialog contexts. 
Example, the query ``\textit{price=moderate}" fetches non-zero rewards for both intents in Table \ref{tab:ca}. Such phenomena can confuse an RL agent, to the extent that it may end up incorrectly learning a single partial query as the best query for many contexts. 
%-- this is because a single partial query could fetch high rewards for multiple dialogs, .

This problem is further exacerbated if some query attributes in $KB$ are correlated. For example (a) in Table \ref{tab:ca}, let us assume that $KB$ has 8 \textit{chinese} restaurants out of which 7 have \textit{moderate} price range, i.e., cuisine and price range are highly correlated. Here, results of the partial query ``cuisine=\textit{chinese}" would contain only one additional restaurant compared to the complete query. As a result, they will receive almost the same reward during training. This could potentially get extreme in certain cases, where presence or absence of an attribute makes no difference, giving little signal to an RL agent. As the number of attributes in an intent increases, the problem can get harder and harder. In our preliminary experiments using MAPO with reward from Equation \ref{eq:reward}, the RL agent often produced partial queries leading to further errors in subsequent dialog.

In our datasets intents can be described by SELECT queries with multiple WHERE clauses; thus, every partial query is a prefix of some complete query (e.g, query 3 in Table \ref{tab:ca}). 
%For example, in Table \ref{tab:ca}, the partial query 3 is a prefix for the complete query 1 in both the examples. 
Upon further analysis of model behavior, we observed that MAPO often maintained both partial prefix queries and complete queries, but still learned a model to output the partial ones.

To understand this surprising observation, we first note that auto-regressive decoders, due to probability multiplication at every decode step, generally prefer \emph{shorter} grammatical sentences (partial prefix query in our case) to longer ones. This is particularly likely to happen when the decoder gets randomly initialized at the start of training. 
A partial query in the buffer that is assigned a high probability makes a higher contribution in the gradient estimation. Moreover, because of correlated attributes, this query may have a fairly high reward. This results in the model believing it to be a good query, and changing the parameters to increase its probability further. The only way RL can break out of this vicious cycle is if the complete query is explored often -- however, MAPO's in-buffer sampling probabilities are proportional to network-assigned probabilities, leading to an ineffective in-buffer exploration. This issue is general, but gets extreme in our problem due to a combination of (1) real-valued (not just 0/1) rewards, (2) prefix queries getting high rewards due to correlated attributes, and (3) prefix queries getting sampled more often due to decoder's bias in favoring shorter queries.

%\newtext{Further investigation showed that the presence of non-zero reward partial queries that are exact prefix of the complete intent query were responsible for MAPO to learn a policy that often produce partial queries. We noticed that the buffer contains one or more partial queries that are a prefix to the complete intent query. For example, in Table \ref{tab:ca}, the partial query 3 is a prefix for the complete query 1 in both the examples. As the policy used is auto-regressive, a randomly initialization will assign the prefix partial query with much higher probability compared to the complete query. The assigned high probability ensures the prefix  queries to have a higher contribution in the gradient estimation. This would result in the partial query being assigned even higher probability. This loop leads to poor exploration of queries inside the buffer, and forces the policy to learn the prefix queries even though the buffer may contain complete queries with much higher reward.}

%To counter this problem, the gradient estimation technique must be able to better explore queries in the buffer. 
In response, our proposed baseline extends MAPO to maintain multiple buffers, so that complete intent queries with high rewards can be prioritized over other queries during gradient estimation. We use two buffers with MAPO: a buffer $\mathcal{B}_h$ to store all queries with the highest reward for a dialog context and a buffer $\mathcal{B}_o$ to store all other queries whose rewards are non-zero and less than the highest. The expected reward is now computed as:
\begin{align}
\mathcal{O}_{ER} = \sum_{\textbf{a} \in \mathcal{B}_h} \pi_{\theta}(\textbf{a}) R(\textbf{a}) + \sum_{\textbf{a} \in \mathcal{B}_o} \pi_{\theta}(\textbf{a}) R(\textbf{a}) \notag \\ + \!\!\sum_{\textbf{a} \in \mathcal{A} - (\mathcal{B}_h \cup \mathcal{B}_o) } \pi_{\theta}(\textbf{a}) R(\textbf{a}) \notag \\
= \pi_{\mathcal{B}_h} \mathbb{E}_{\textbf{a} \sim \pi_{\theta}^{h+}(\textbf{a})} R(\textbf{a}) + \pi_{\mathcal{B}_o} \mathbb{E}_{\textbf{a} \sim \pi_{\theta}^{o+}(\textbf{a})} R(\textbf{a}) \notag \\ + (1-\pi_{\mathcal{B}_h}-\pi_{\mathcal{B}_o}) \mathbb{E}_{\textbf{a} \sim \pi_{\theta}^{-}(\textbf{a})} R(\textbf{a})
\label{eq:mmapo}
\end{align}
where $\pi_{\theta}^{h+}$, $\pi_{\theta}^{o+}$ and $\pi_{\theta}^{-}$ are the normalized probability distributions of queries in $\mathcal{B}_h$, queries in $\mathcal{B}_o$ and all other queries respectively. $\pi_{\mathcal{B}_h}=\sum_{\textbf{a} \in \mathcal{B}_h} \pi_{\theta}(\textbf{a})$  and $\pi_{\mathcal{B}_o}=\sum_{\textbf{a} \in \mathcal{B}_o} \pi_{\theta}(\textbf{a})$ are the total probabilities assigned by the policy   $\pi_{\theta}$ of all queries in $\mathcal{B}_h$ and $\mathcal{B}_o$ respectively. As the complete intent queries mostly have the highest rewards for a given context, they are placed in $\mathcal{B}_h$, and as the partial prefix queries usually have rewards less than the complete queries, they are placed in the other buffer $\mathcal{B}_o$. To ensure all non-zero reward queries are explored, each buffer is made to contribute to the gradient estimation. To ensure each buffer contributes significantly to the gradient estimation, we clip $\pi_{\mathcal{B}_h}$ using $\text{max}(\pi_{\mathcal{B}_h}, \alpha_h)$ and $\pi_{\mathcal{B}_o}$ using $\text{min}(\text{max}((1-\pi^c_{\mathcal{B}_h})\alpha_o, \pi_{\mathcal{B}_o}), (1-\pi^c_{\mathcal{B}_h}))$.
\begin{comment}
\begin{align}
\pi^c_{\mathcal{B}_h} &= \text{max}(\pi_{\mathcal{B}_h}, \alpha_h) \notag \\
\pi^c_{\mathcal{B}_o} &= \text{min}(\text{max}((1-\pi^c_{\mathcal{B}_h})\alpha_o, \pi_{\mathcal{B}_o}), (1-\pi^c_{\mathcal{B}_h})) \notag 
\end{align}
\end{comment}
%where 
$\alpha_h$ and $\alpha_o$ are hyper-parameters whose value can be between 0 and 1. $\alpha_h$ ensures the highest reward buffer gets assigned a certain weight during gradient estimation. $\alpha_o$ ensures the  queries in $\mathcal{B}_o$ are assigned at least a certain fraction of probability mass unused by the queries in $\mathcal{B}_h$. The estimator is biased when the training starts, and can becomes unbiased when $\pi_{\mathcal{B}_h} > \alpha_h$ and $\pi_{\mathcal{B}_o} > (1-\pi_{\mathcal{B}_h}))\alpha_o$. Based on the clipped probabilities, the gradients are estimated as:
\begin{align}
\nabla_{\theta} \mathcal{O}^{c}_{ER}
= \pi^c_{\mathcal{B}_h} \mathbb{E}_{\textbf{a} \sim \pi_{\theta}^{h+}(\textbf{a})} \nabla_{\theta} \text{log} \pi_{\theta}(\textbf{a}) R(\textbf{a}) \notag \\ + \pi^c_{\mathcal{B}_o} \mathbb{E}_{\textbf{a} \sim \pi_{\theta}^{o+}(\textbf{a})} \nabla_{\theta} \text{log} \pi_{\theta}(\textbf{a}) R(\textbf{a}) \notag \\ + (1-\pi^{c}_{\mathcal{B}_h}-\pi^{c}_{\mathcal{B}_o}) \mathbb{E}_{\textbf{a} \sim \pi_{\theta}^{-}(\textbf{a})} \nabla_{\theta} \text{log} \pi_{\theta}(\textbf{a}) R(\textbf{a})
\label{eq:gemmapo}
\end{align}
Finally, if a new query is found that has a higher reward than queries in $\mathcal{B}_h$, then the buffers are updated: that query is added to $\mathcal{B}_h$, and all other queries in that buffer are removed and added to $\mathcal{B}_o$. The proposed approach can be used for estimating the policy gradients for any deterministic environments with discrete actions and non-binary rewards.

\begin{figure*}[ht]
\centering
\includegraphics[width=\textwidth]{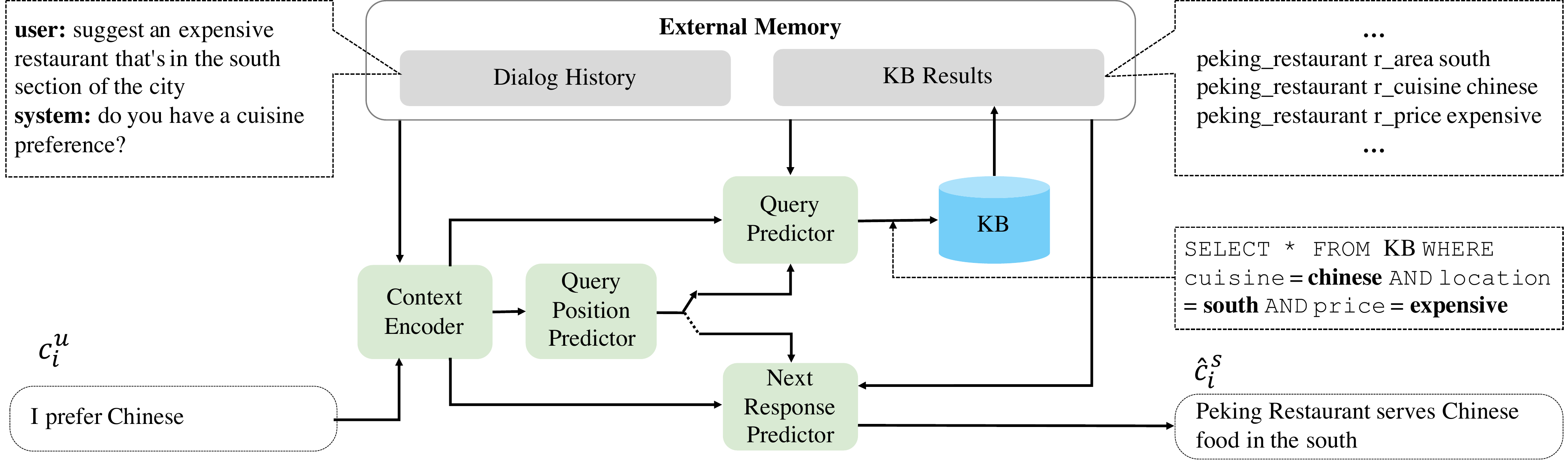}
\caption{The architecture of the uTOD system.}
\label{fig:archi}
\end{figure*}

\subsection{uTOD System}
\label{ssec:utod}

We now describe the remaining parts of the uTOD system that learns to predict the next response using unannotated dialogs. The system has three main components: (1) query position predictor, (2) query predictor (as described in previous section) and (3) next response predictor. The system architecture is shown in Figure \ref{fig:archi}.
The position predictor is a binary classifier that decides whether the KB query is to be predicted at the given turn or not. The query predictor generates a KB query at the turn predicted by the position predictor. Finally, the next response predictor takes the dialog context along with the query results (if a KB query has been made) and predicts the next system response.

%\vspace*{0.5ex}
%\noindent {\bf Query Position Predictor:}
In order to train the position predictor, each train dialog must be annotated with the turn at which the KB query is to be made. As there are no gold labels available, we heuristically provide this supervision. We identify the turn ($\tilde{q}$) at which the agent response contains a KB entity that was never seen in the dialog context. We mark $\tilde{q}$ as the heuristic label for training position predictor.\footnote{This heuristic labeling matched ~80\% of gold labels in our datasets} For example, in Figure~\ref{fig:examples}, $t=2$ is the turn at which the agent response contains a KB entity \textit{Peking restaurant} that was never used in the dialog context. 

The position predictor takes the dialog context as input and encodes it using a multi-hop memory encoder, as described in Section \ref{ssec:query-predictor}. The encoder output is then passed through a linear layer followed by sigmoid function to generate the probability of the binary label.  At training time, KB query predictor learns to generate KB queries at the turns ($\tilde{q}$) annotated by the heuristic. During test time, a query is generated at the turn ($\hat{q}$) predicted by the position predictor. 

The query predictor can be seen as annotating all the train dialogs with KB queries. Thus, after this annotation, any existing end-to-end TOD system can be used for next response prediction. For our experiments, we use BoSsNet as the underlying response predictor \cite{raghu2019disentangling}. \textcolor{black}{BoSsNet takes as input  (i) the dialog context and (ii) the ground truth KB query results (during both train and test) to generate the response. As we assume the ground truth KB queries are unavailable, we use the results of the predicted query results instead.}

Since our system uses a pipeline of components, it is trained in a
 training curriculum, which is divided into three phases.
In \textit{phase-1}, the query predictor is trained. Once each train dialog is annotated with KB queries, the next response predictor is trained in \textit{phase-2}. Finally, in \textit{phase-3} the position predictor is trained. We train the position predictor after the response predictor as we observed that initializing the position predictor encoder with the weights of the response predictor encoder yields better position prediction performance.

\section{Experimental Setup}
We now describe the datasets used, the comparison algorithms and the evaluation metrics for our task. We release all software and datasets from our experimental setup for further research.

\subsection{Datasets} 
We perform experiments on two task-oriented dialog datasets from the restaurant researvation domain: CamRest~\cite{wenEMNLP2016} and DSTC2~\cite{henderson2014second}. \textcolor{black}{CamRest676 is a human-human dialog dataset with having just one KB query annotated in them. DSTC2 is a human-bot dialog dataset. We filtered the dialogs from this dataset which had more than one KB query.}
Table \ref{tab:statistics} summarizes the statistics of the two datasets. 
%All the dialogs in the dataset have just one KB query annotated in them.
As CamRest and DSTC2 are originally designed for dialog state tracking, we use the versions that are suitable for end-to-end learning, made available by \newcite{raghu2019disentangling} and \newcite{BordesW16}, respectively. Both these datasets have KB query annotations. We remove these annotations from the training dialogs to create datasets for our task. 
We use these annotations for evaluating the performance of various algorithms.

%CamRest is a subset of the MultiWOZ dataset \cite{budzianowski2018multiwoz}, however, many domains in MultiWOZ are not suitable for evaluating our problem, since they do not have associated KBs.

\begin{table}[ht]
\centering
\footnotesize
\begin{tabular}{l c c}
\toprule
& CamRest & DSTC2\\
\midrule
Train Dialogs & 406 & 1279 \\
Val Dialogs & 135 & 324 \\
Test Dialogs & 135 & 1051 \\
Avg. no. of turns & 4.06 & 7.94 \\
Rows/Fields in KB  & 110/8 & 108/8 \\
Vocab Size & 1215 & 958 \\
\bottomrule
\end{tabular}
\caption{Statistics of CamRest and DSTC2 datasets.} 
\label{tab:statistics}
\end{table}

\subsection{Comparison Baselines}

%We study the problem of predicting KB query annotations with the following training algorithms\footnote{We will release the code for all algorithms and datasets for further use.}:

We compare various gradient estimation techniques to train the policy network (defined in Section \ref{ssec:query-predictor})  for the task of KB query prediction as follows.

\vspace{0.5ex}
\noindent{\bf REINFORCE}~\cite{williams1992simple}: uses on-policy samples to estimate the gradient as in Equation \ref{eq:reinforce}, with the reward function defined in Equation \ref{eq:reward}.

\vspace{0.5ex}
\noindent{\bf BS-REINFORCE}~\cite{guu2017language}: \textcolor{black}{uses all the queries generated by beam search to estimate the gradients. The reward for each query is a product of the reward defined in Equation \ref{eq:reward} and the likelihood of the query in the current policy.}

\vspace{0.5ex}
\noindent{\bf RBS-REINFORCE}~\cite{guu2017language}: \textcolor{black}{uses all the queries generated by $\epsilon$-greedy randomized beam search (RBS) to estimate the gradients. At every time step, RBS samples a random continuation as opposed to a highest scoring continuation with a probability $\epsilon$.}

\vspace{0.5ex}
\noindent{\bf MAPO}~\cite{liang2018memory}: uses on-policy samples and a buffer of non-zero reward queries to estimate the gradients. It uses the reward function defined in Equation \ref{eq:reward}.

\vspace{0.5ex}
\noindent{\bf m$\mathcal{B}$-MAPO}: the proposed approach, multi buffer MAPO, which uses a buffer with highest reward queries and a buffer with other non-zero reward queries to estimate the gradients. This algorithm uses the reward function defined in Equation \ref{eq:reward}.

\vspace{0.5ex}
\noindent{\bf SL}: uses the gold KB queries as direct supervision. We train the policy network proposed in Section \ref{ssec:query-predictor} with cross entropy loss.

\vspace{0.5ex}
\noindent{\bf SL+RL}: we combine weak supervision using RL as a secondary source of knowledge on top of supervised cross entropy loss for better training. This is analogous to the benefits of additional prior knowledge (e.g., as constraints) on top of supervision for small datasets \cite{nandwani-nips19}. In our experiments, \emph{total loss = Cross-Entropy }$- \lambda.\mathcal{O}_{ER}$. Equation \ref{eq:mmapo} is used for computing $\mathcal{O}_{ER}$.

%In particular, they use the beam search guided by the SQL grammar which generates only syntactically correct SQL queries.
%\newtext{Uses the gold KB queries as direct supervision. We train the policy network proposed in Section \ref{ssec:query-predictor} with a linear combination of cross entropy loss and m$\mathcal{B}$-MAPO objective defined in Equation \ref{eq:mmapo}. The cross entropy loss and the  m$\mathcal{B}$-MAPO objective have weights of one and $\gamma$ respectively. $\gamma$ is a hyperparameter which can take a value between 0 and 1.}

%It serves as a reference for the other approaches.

%To study how the predicted annotations affect the end goal of learning TOD systems, we use the predicted queries to train, an end-to-end TOD system~\cite{raghu2019disentangling} and analyze the performance.

We also compare the full $\text{uTOD}$ dialog engines, built using these RL approaches for query prediction -- we name them $\text{uTOD}^{\text{REINFORCE}}$,  $\text{uTOD}^{\text{MAPO}}$ and $\text{uTOD}^{\text{mMAPO}}$. We also compare these to a fully supervised TOD system that uses KB query annotations during training ($\text{aTOD}$). We report two variants of the $\text{aTOD}$ system based on the supervised query predictors used: $\text{aTOD}^{\text{SL}}$ and $\text{aTOD}^{\text{SL+RL}}$.

%Our primary goal is to build a $\text{uTOD}$ system, a TOD system that learns using unannotated dialogs. We compare its performance to $\text{aTOD}$ system that learns using annotated dialogs. To refer to a uTOD system whose query predictor is trained using a specific algorithm, we use the algorithm name as a superscript. For example, a uTOD system whose query predictor is trained using MAPO is referred to as $\text{uTOD}^\text{MAPO}$.

\subsection{Evaluation Metrics}
As we have gold annotations, we evaluate the KB query predictor, position predictor separately, in addition to the overall TOD system. KB query predictors are evaluated based on accuracy -- the fraction of dialogs where the gold KB queries are predicted. We also report two other metrics: total reward and the PIQ ratio -- the fraction of dialogs where a partial intent query was outputted, i.e., the predicted query captured only a subset of gold query attributes.
%The algorithms used for training the query predictor are evaluated based on the accuracy with which they predict the gold KB queries. Since both the datasets have the KB queries annotated, we use them to compute the accuracy.

A query position predictor's performance is also measured using accuracy. The predictor is considered correct only if it predicts 1 at the turn at which query is made and 0 for all turns before that. We also compute \emph{turn difference} as the absolute difference between the turn at which the classifier predicts true and the turn corresponding to the gold label in the annotated dialog. The smaller the average turn difference, the better is the classifier.

The next response prediction is evaluated based on its ability to match the gold responses at every turn. We use standard metrics of BLEU \cite{papineni2002bleu} and entity F1 to measure the similarity between predicted and gold responses. Entity F1 is the average of F1 scores computed for each response. We also report Entity F1 $KB $ for entities that can only be copied from the KB results, to emphasize importance of using query results in subsequent dialog.

We note that our setup for evaluating dialog engines looks similar to supervised TOD systems, but has a subtle but important difference. In standard supervised dialog evaluations, for a given turn, the full dialog context is provided and the system is evaluated on correctly predicting the next utterance. Thus, even if the system made an incorrect query at an earlier turn, the subsequent turns will be shown correct query and correct KB query results.

However, since our goal is to assess the response prediction without query annotation, we remove such annotation from the test dialogs also. That is, for the test dialogs subsequent to making the query, all responses are generated based on the \emph{predicted} KB queries at predicted position $\hat{q}$ and their query results. This makes the evaluation much more realistic, but also quite challenging for current dialog engines. This is the key reason why aTOD performance in Table \ref{tab:tod-study} is much lower than results reported in the BossNet paper \cite{raghu2019disentangling}.

%The predicted query annotations are then used to train the downstream TOD system. These systems are evaluated based on their ability to generate valid responses. We use that standard metrics, BLEU \cite{papineni2002bleu} and entity F1, to measure the similarity between predicted and gold responses. Entity F1 is the average of F1 scores computed for each response.
%We also collect human judgements to compare the quality of the responses generated.

%The query position predictor is a binary classifier that indicates whether a KB query should be made at the given turn or not. During test time, the classifier should predict true only when the user has specified all query attributes and false for all turns before that. A sequence of correct predictions is necessary for the QP predictor to be effective for a single dialog. To capture this behaviour, we compute \emph{dialog accuracy}, where the classifier is correct only if it predicts one at the turn at which query is to be made and zero for everything before. We also compute \emph{turn difference} as the absolute difference between the turn in the dialog at which the classifier predicts true and the turn corresponding to the gold label in the annotated dialog. The smaller the average turn difference, the better is the classifier.

We perform two human evaluation experiments to compare (1) \textit{informativeness} - the ability to effectively use the results to generate responses and convey the information requested by the user, and (2) \textit{grammar} --  ability to generate grammatically correct and fluent responses. Both the dimensions were annotated on a scale of (0–2). As the primary focus of our work is to evaluate the ability of a TOD system to effectively use the annotated query results, we only collect judgements for responses that occur {\em after} the KB query. We sampled 100 random dialog-context from CamRest dataset and collected judgements from 2 judges for 4 systems, namely $\text{uTOD}^\text{REINFORCE}$, $\text{uTOD}^\text{MAPO}$, $\text{uTOD}^\text{\sys}$ and $\text{aTOD}$. We collected a total of 1600 labels from the judges.

\subsection{Implementation Details}
We implemented our system using TensorFlow \cite{abadi2016tensorflow}. 
%Word embeddings and weights were initialized using a standard normal distribution with mean $0$ and variance $0.01$. We trained the network using an Adam optimizer \cite{kingma2014adam} and applied gradient clipping with a clip-value of 40.
 We identify hyper-parameters based on the evaluation of the held-out validation sets.  We sample word embedding, hidden layer, and cell sizes ($es$) from $\{32, 64, 128, 256, 512\}$, learning rates ($lr$) from $\{10^{-3}, 25\times10^{-4}, 5\times10^{-4}, 10^{-4}\}$, $\alpha_o$ and $\lambda$ from increments of 0.1 between [0.1, 0,9], $\epsilon$ from $\{0.05, 0.1, 0.15, 0.2\}$, and $\alpha_h$ from increments of 0.1 between [0.5, 0,9]. The hyper-parameters that $(\alpha_h, \alpha_o, \lambda, \epsilon, es, lr)$ achieved the best validation rewards were $(0.5, 0.1, 0, 0.15, 256, 5\times10^{-4})$ and $(0.6, 0.1, 0.1, 0.15, 256, 25\times10^{-4})$ for DSTC2 and CamRest respectively. As our response predictor is BossNet, we used the best performing hyper-parameters reported by \newcite{raghu2019disentangling} for each dataset. Total accumulated validation rewards is used as a early stopping criteria for training the query predictor and BLEU for the training the response predictor. 

\section{Experiments}
Our experiments evaluate three research questions:
% \todo{Is this ordering correct?}
\begin{compactenum}
    \item \textit{Query Predictor Performance:} How does the performance of \sys\ compare to other gradient estimation techniques? 
    %\item \textit{Ablation Study:} What is the performance gain from  each feature in \sys?
    \item \textit{Query Position Predictor Performance:} How does the proposed position predictor perform on the two datasets? 
    \item \textit{Next Response Predictor Performance:} How do responses from TOD systems trained with unannotated dialogs compare to the ones trained with KB query annotated dialogs?
\end{compactenum}

\begin{table*}[ht]
\centering
\footnotesize
\begin{tabular}{l|c c|c c|c c}
\toprule
& \multicolumn{2}{c|}{\textbf{Accuracy}} & \multicolumn{2}{c|}{\textbf{PIQ Ratio}} & \multicolumn{2}{c}{\textbf{Total Test Rewards}} \\
\cmidrule{2-7}
& \textbf{DSTC2} & \textbf{CamRest} & \textbf{DSTC2} & \textbf{CamRest} & \textbf{DSTC2} & \textbf{CamRest} \\
\midrule
REINFORCE & 0.00 & 0.00 & 0.00 & 0.00 & 0.00 & 0.00 \\
BS-REINFORCE & 0.00 & 0.00 & 0.00 & 0.004$\pm$0.008 & 0.00 & 0.01$\pm$0.03 \\
RBS-REINFORCE & 0.00 & 0.002$\pm$0.005 & 0.00 & 0.02$\pm$0.05 & 0.00 & 0.24$\pm$0.26 \\
MAPO  & 0.25$\pm$0.01 & 0.10$\pm$0.01 & 0.61$\pm$0.01  & 0.54$\pm$0.05  & 77.13$\pm$3.6 & 11.8$\pm$0.5 \\
\sys\ & \textbf{0.68$\pm$0.03} & \textbf{0.62$\pm$0.03} &  \textbf{0.06$\pm$0.03} & \textbf{0.09$\pm$0.02} & \textbf{169.98$\pm$7.6} & \textbf{23.4$\pm$1.1} \\
\midrule
\midrule

SL & 0.78$\pm$0.02 & 0.59$\pm$0.06 & 0.09$\pm$0.02  & 0.09$\pm$0.03  & 175.9$\pm$2.7 &  21.7$\pm$1.4 \\

SL+RL & 0.78$\pm$0.02 & 0.66$\pm$0.04 & 0.09$\pm$0.02  & 0.09$\pm$0.03  & 175.9$\pm$2.7 &  23.7$\pm$1.3 \\
\bottomrule
\end{tabular}
\caption{Accuracy of KB query prediction of \sys\ and other algorithms on CamRest and DSTC2 on 10 runs. Partial intent query (PIQ) ratio is the fraction of partial intent queries predicted.}
\label{tab:performace-study}
\end{table*}

\subsection{Query Predictor Performance}

To measure the query predictor performance, we generate the KB queries at the gold position during train and test. We only use the gold queries during test to measure the performance. Table~\ref{tab:performace-study} reports the KB query prediction accuracy, PIQ ratio and total test rewards achieved by various gradient estimation techniques. 
%PIQ ratio is the fraction of predicted queries in the test set that captured only subset of the gold query attributes. 
%For a fair comparison, the rewards reported are computed using Equation \ref{eq:reward}.
% describe why our approach is better than MAPO
\sys\ significantly outperforms  MAPO on both datasets. The performance gain comes from \sys's ability to address correlated attributes in $KB$ and the frequent sampling of highest reward queries from the buffer to prevent the policy from learning common partial intent queries. Compared to MAPO, \sys\ reduces the PIQ ratio by 55\% on DSTC2 and 54\% on CamRest, and achieves considerably higher rewards.

% describe why REINFORCE is bad
The failure of REINFORCE highlights that using just the on-policy samples to estimate policy gradients is inadequate for exploring large combinatorial search spaces with sparse rewards. Both MAPO and \sys\ use queries in the buffer, which are explored using systematic search. These guide the policy towards the parts of the search space that are likely to yield non-zero rewards.

% describe why our approach does better than Supervised in CamRest
% 22 percent, 8 and 6 percent.
We notice that, surprisingly, \sys\ achieves slightly better accuracy than even the supervised (SL) baseline on CamRest dataset. Further analysis reveals that on this dataset, supervised learner achieves a train accuracy of 95\%, while \sys\ achieves only 75\% (but higher test accuracy). This suggests that the supervised learner is overfitting, which is conceivable since CamRest is a relatively small dataset (406 train dialogs, see Table \ref{tab:statistics}). Because \sys\ is learning with weak supervision, it is solving a much harder problem, which makes it harder to overfit. Moreover, sometimes \sys\ may simply learn partial intent queries if they generalize better. This added flexibility helps avoid overfitting in the small dataset.
%\todo{M: I am not sure if i buy this argument. why would flexibility reduce overfitting in small data.} 
Our error analysis reveals that 25\% of queries generated by supervised learner had non-entity words as values in the WHERE clause. For example,``please" was predicted as a cuisine. This shows the model has learned to pick up spurious signals to predict certain attributes. In contrast, only 11-13\% of the queries predicted by MAPO and \sys\ exhibited this problem. To prevent the supervised learner from overfitting, SL+RL combines the weak supervision using RL as a secondary source of knowledge on top of cross entropy loss. This increased the accuracy by 7 points and percent of queries with non-entity words reduced from 25\% to 11\%.

%On CamRest, \sys\ achieves better accuracy than the supervised approach. The inferior performance of supervised approach is a result of overfitting caused by a combination of the dataset size of the dataset and objective function used. The cross-entropy objective function maximizes the likelihood of the gold query. When the dataset is small, the network could pick up spurious signals and hence fails to generalize well. On the other hand, the RL objective function is more relaxed and provides flexibility for the model to learn just the partial intent queries which may lead to better generalization. Such an objective function helps avoiding over-fitting, when the dataset is small. On Camrest, the supervised learner achieved an average train accuracy of 95\%, while \sys\ achieved only 75\%. The performance on validation for both the dataset were around 62\% and 65\% respectively. We noticed that 17\% of the queries predicted by the supervised learner had non-entity words as values in the WHERE clauses. For example,``please" was predicted as a cuisine. This shows the model has learned to pick up spurious signals to predict certain attributes. Only 6\% of the queries predicted by MAPO and 8\% by \sys\ exhibited this problem.

% describe the gap between our approach and Supervised in DSTC2
On the other hand, compared to supervised learner, \sys\ is 10 accuracy points lower on DSTC2. The difference in performance can be attributed to two factors. First, DSTC2 is a larger dataset, which enables supervised learner to train well. Second, in this dataset, there are 12\% dialogs where overconstrained queries (those that have even more attributes than the gold) fetch better rewards than the gold. For instance, in Table \ref{tab:ca} example (a), say the query ``cuisine=\textit{chinese} AND price=\textit{moderate} AND location=\textit{west}" fetches all the entities mentioned in subsequent dialog. Even though it has one additional attribute compared to the actual user intent (or gold query), it will likely contain fewer unused entities than the gold query. Thus \sys\ will assign a higher reward to this overconstrained query, and will encourage the training to output this. This confuses \sys\ leading to a significant performance gap from supervised learner.

%Compared to supervised approach, \sys\ is almost 14 points lower on DSTC2. The difference in accuracy is due to the design of the reward function. Our reward function encourages the agent to predict queries that returns all subsequent entities, and penalizes it for retrieving more than required entities. While the reward function approximation works for most cases, there are cases where certain queries are assigned higher reward compared to the gold query. In example (a) in Table \ref{tab:ca}, let us assume the query ``cuisine=\textit{chinese} AND price=\textit{moderate} AND location=\textit{west}" fetches all the relevant entities. Even though it has one additional attribute compared to the actual user intent (or gold query), it contains fewer unused entities than the gold query. Thus \sys\ will assign a higher reward to this query. In DSTC2, there were quite a few of these cases. Almost 11\% of the queries predicted by \sys\ contained more attributes than their corresponding gold queries.

\begin{figure}
\centering
\includegraphics[width=0.48\textwidth]{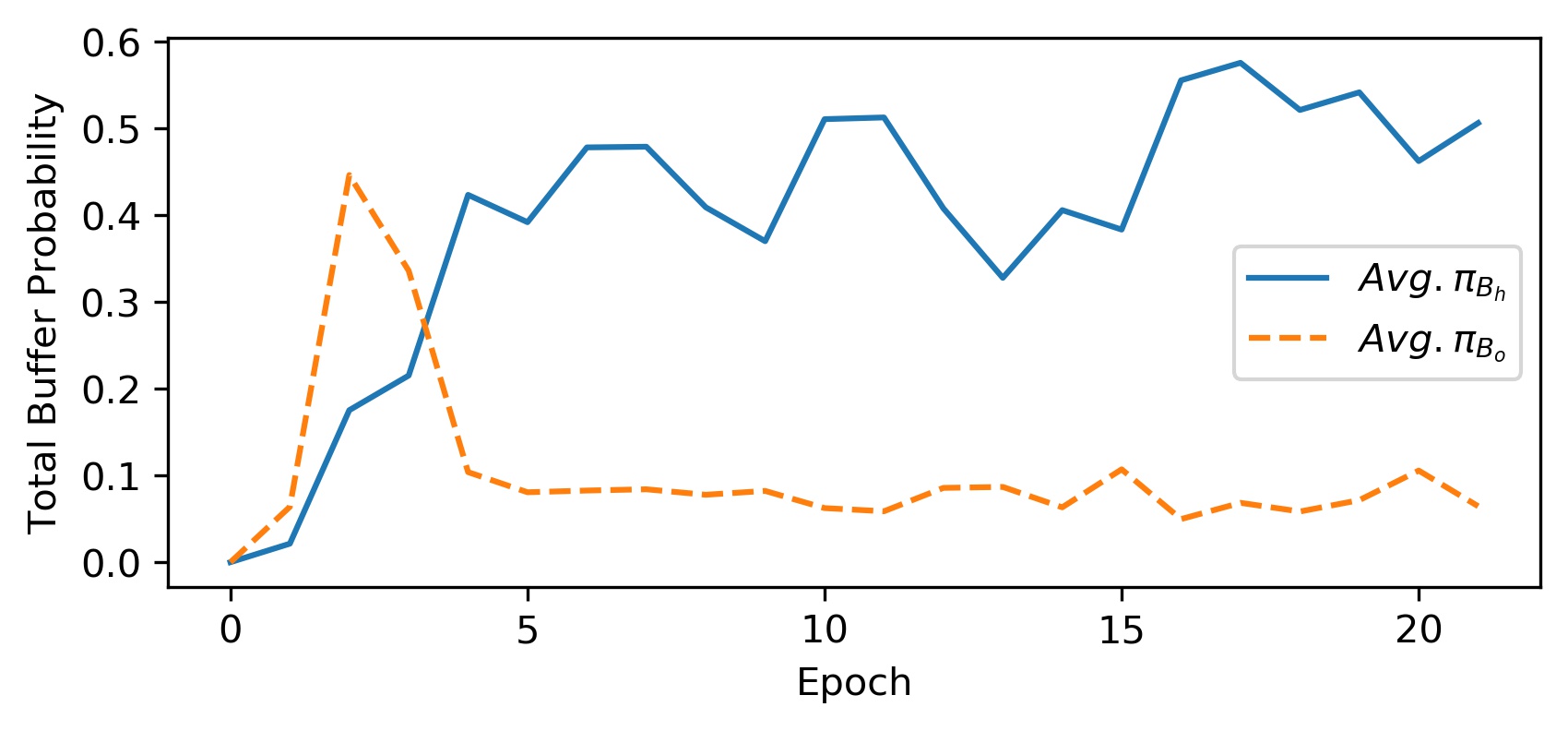}
\caption{Dynamics of total buffer probabilities $\pi_{\mathcal{B}_h}$ and $\pi_{\mathcal{B}_o}$ while training on DSTC2}
\label{fig:dynamics}
\end{figure}

\noindent \textbf{Dynamics of the total buffer probabilities:}
$\pi_{\mathcal{B}_h}$ and $\pi_{\mathcal{B}_o}$ are the sum of probabilities of all the queries in buffers $\mathcal{B}_h$ and $\mathcal{B}_o$ respectively. We now discuss the dynamics of  $\pi_{\mathcal{B}_h}$ and $\pi_{\mathcal{B}_o}$ during training. We define average $\pi_{\mathcal{B}_h}$ and average $\pi_{\mathcal{B}_o}$ as the average of total buffer probabilities across all the examples in train. Figure \ref{fig:dynamics} shows the average $\pi_{\mathcal{B}_h}$ and average $\pi_{\mathcal{B}_o}$ after each train epoch on DSTC2. Queries in ${B}_o$ are typically shorter (partial queries) compared to the queries in ${B}_h$ and so they get assigned a higher probabilities when the policy is randomly initialized. Hence during initial epochs average $\pi_{\mathcal{B}_o}$ is higher than average $\pi_{\mathcal{B}_h}$. But as the training proceeds, the policy learns to assign higher probability to the (longer) queries in ${B}_h$ as a result of clipping the buffer probabilities defined by $\alpha_h$. The gradients are biased towards the queries in ${B}_h$ during the first few epochs, but as the policy converges we can see that the average $\pi_{\mathcal{B}_h}$ reaches close to $\alpha_h$ (0.5 for DSTC2) making the gradient estimates unbiased.
\begin{comment}
\subsection{Ablation Study}

We perform an ablation study where we start with MAPO and incrementally add the modifications to counter correlated attributes (CA) and dialog variations (NORM). Table~\ref{tab:ablation-study} reports the KB query prediction accuracy of variants on CamRest and DSTC2 datasets. MAPO+CA uses only the highest reward queries in the buffer with the reward function given by Equation \ref{eq:ca-reward}. MAPO+CA+NORM is same as \sys. It uses the normalized reward defined by Equation \ref{eq:final-reward}. We find that both model components add value to the final performance gain.\todo{M: any intuition why norm doesnt add value in camrest?}

\begin{table}[ht]
\centering
\small
\begin{tabular}{l c c}
\toprule
             & DSTC2     & CamRest   \\
\midrule
MAPO  & 0.41$\pm$0.16 & 0.44$\pm$0.08 \\
\hspace{2mm} +CA & 0.54$\pm$0.04 & 0.62$\pm$0.04 \\
\hspace{2mm} +CA+NORM & \textbf{0.64$\pm$0.08} & \textbf{0.63$\pm$0.03} \\
\bottomrule
\end{tabular}
\caption{Query prediction accuracy of \sys\ without modification to counter each issue.}
\label{tab:ablation-study}
\end{table}
\end{comment}

\begin{table}[ht]
\centering
\small
 \begin{tabular}{l|c c | c c}
\toprule
%& \multicolumn{2}{c|}{} & \multicolumn{2}{c}{\textbf{Avg. Turn}} \\
%& \multicolumn{2}{c|}{\textbf{Accuracy}} & \multicolumn{2}{c}{\textbf{Difference}}
& \multicolumn{2}{c|}{\textbf{Accuracy}} & \multicolumn{2}{c|}{\textbf{ATD}}
\\
\cmidrule{2-5}
& w/o PT & w/ PT & w/o PT & w/ PT \\
\midrule
CamRest & 0.47 & \textbf{0.54} & 0.72 & \textbf{0.56} \\
DSTC2 & 0.21 & \textbf{0.28} & 2.26 & \textbf{2.20} \\
\bottomrule
\end{tabular}
\caption{Position predictor performance with and without pre-training (PT). ATD = average turn difference.} 
\label{tab:classifier}
\end{table}

\begin{table*}[ht]
\centering
\small
\begin{tabular}{l | c c c | c c c}
\toprule
& \multicolumn{3}{c|}{\textbf{DSTC2}} & \multicolumn{3}{c}{\textbf{CamRest}} \\
\cmidrule{2-7}
& \textbf{Ent. F1} & \textbf{Ent. F1} & \textbf{BLEU} & \textbf{Ent. F1} & \textbf{Ent. F1} & \textbf{BLEU} \\
& KB & All & & KB & All & \\
\midrule
$\text{uTOD}^{\text{REINFORCE}}$ & 0.11$\pm$0.01  & 0.36$\pm$0.02 & \textbf{51.52$\pm$0.91} &
0.02$\pm$0.01 & 0.34$\pm$0.02 & \textbf{15.09$\pm$0.56} \\

%\midrule

%$\text{uTOD}^{\text{MAPO}}$ & 0.17$\pm$0.01 & 0.35$\pm$0.02 & 45.79$\pm$0.80 &     
% 0.17$\pm$0.01 & 0.34$\pm$0.03 & 13.40$\pm$0.74 \\
$\text{uTOD}^{\text{MAPO}}$ & 0.14$\pm$0.01 & 0.36$\pm$0.01 & 46.27$\pm$1.63 &     
 0.14$\pm$0.01 & 0.31$\pm$0.03 & 12.90$\pm$0.75 \\

%$\text{uTOD}^{\text{\sys}}$ & 0.23$\pm$0.02 & 0.37$\pm$0.02 & 46.33$\pm$0.75 &      
%0.22$\pm$0.01 & 0.35$\pm$0.03 & 13.69$\pm$0.77 \\
$\text{uTOD}^{\text{\sys}}$ & \textbf{0.21$\pm$0.02} & \textbf{0.38$\pm$0.02} & 47.52$\pm$1.27 &     
 \textbf{0.23$\pm$0.02} & \textbf{0.35$\pm$0.02} & 13.11$\pm$0.86 \\
 
\midrule
$\text{aTOD}^{\text{SL}}$ & 0.24$\pm$0.02 & 0.41$\pm$0.02 & 48.35$\pm$1.58 &
 0.25$\pm$0.03 & 0.36$\pm$0.02 & 13.80$\pm$1.04 \\
$\text{aTOD}^{\text{SL+RL}}$ & 0.24$\pm$0.02 & 0.41$\pm$0.02 & 48.35$\pm$1.58 &
 0.29$\pm$0.03 & 0.41$\pm$0.02 & 14.68$\pm$0.85 \\
\bottomrule
\end{tabular}
\caption{Performance of various uTOD systems and aTOD system on 10 runs.}
\label{tab:tod-study}
\end{table*}

\subsection{Query Position Predictor Performance}
%Query point predictor predicts the turn at which the KB query is to be generated. 
To study the performance of the query position predictor, we use the gold positions to train the classifier and to measure the performance during test. 
Table~\ref{tab:classifier} shows the accuracy and the average turn differences for the task of query position prediction. The predictor is evaluated in two settings: one initialized with random weights (w/o PT) and one with the encoder network pre-trained for the task of response prediction (w/ PT). The response prediction tasks helps the network identify parts of dialog context that are crucial for generating the response.  This additional information helps improve the accuracy by 7 points on both datasets.

The overall performance for CamRest is acceptable, since average turn difference is less than 1. The errors are due to natural variations in the policy followed by the agent (human in the Wizard of Oz study) -- sometimes the agent fires the query before all possible attributes are specified by the user, and at other times, the agent requests for missing attributes and only then fires a query. However,  performance in DSTC2 is somewhat limited.  In addition to natural agent policy variations, there are artifacts of speech data, because DSTC2 are speech transcripts of human-bot conversations. For exmaple, the bot often re-confirms an already specified attribute to reduce speech recognition errors. As DSTC2 dataset uses transcripts and not actual speech, this re-affirmation is redundant, but is still present. This confuses the model since it has to decide between whether to make a query or request a re-affirmation.

%The errors on CamRest are due to the variations in the policies followed by a human agent. In some cases, the agent fires a KB query even when the user specifies only a subset of all possible query attributes. While, in other cases the agent requests for the missing attributes and once all attributes are specified, the KB query is made. The dialogs in DSTC2 are speech transcripts of human-bot conversations. The errors inb DSTC2 in addition to the variation in agent policy also has variations introduced by the speech-to-text converter. For example, in some cases the bot re-confirms a specified attribute to reduce speech recognition errors. As we use the transcripts and not speech, the predictor finds it hard to decide if a query can be made or should a re-confirmation be done.

\subsection{TOD System Performance}

To study the TOD system performance, we neither use the gold positions nor the gold queries. We use the heuristic defined in Section \ref{ssec:utod} to label the positions. These labelled positions are used to train the position predictor and as the position at which the query predictor generates the query during train. At test time, the query predictor generates the query at the predicted position. 

We study the performance of uTOD systems (next response prediction) trained using the dialogs whose queries were predicted by various query prediction algorithms. For this evaluation, test dialogs are first divided into (context, response) pairs. The system predicts a response for each turn and the predicted response is compared with the gold response. When predicting responses in the subsequent dialog, the results of the predicted query are appended to the context. 
For the aTOD system, results from gold queries are used during train, and the results of predicted queries are used during test. %While for uTOD systems, results of the query predicted by the RL based KB query predictor is used during both train and test. 

Entities in responses can be divided into two types: context entities and KB entities. Context entities are present in the dialog context. For example, in Figure \ref{fig:examples}, agent utterance in turn $2$ has two context entities (``moderate'' and ``south'') and one KB entity (``Peking restaurant''). For the response to contain the correct KB entity, all of position predictor,  query predictor and next response predictor must work together. To assess this, we report \textit{KB Entity F1}, which judges the match between gold and predicted KB entities used in the utterance.

%measures the ability of a TOD system to formulate an appropriate query and infer the entity requested by the user from the results. 

Table \ref{tab:tod-study} shows the performance of various TOD systems on two datasets. We see that the uTOD system trained using \sys\ is only a few points lower than aTOD system. This underscores the value of our gradient computation scheme. \sys\ is significantly better than MAPO -- our analysis reveals that MAPO frequently returns partial queries which have a larger set of results -- this confuses the response predictor, reducing KB Entity F1 substantially.

%The predicted query should return the precise information from the KB for the TOD system to learn efficiently. As MAPO returns more than required entities by predicting partial queries, it becomes hard for the TOD system to infer the necessary information from the large, noisy result set. This results in lower KB entity f1 score compared to \sys. 

Most entities predicted by REINFORCE are just context entities copied from the context and their contribution dominates the entity F1 (all) score. As REINFORCE fails completely in generating correct queries,  the response predictor is forced to memorize the KB entities rather than inferring them from the query results.  This results in very low KB entity F1. As the dialog context often has no query results, the system's only objective becomes generation of good language. Due to this, it achieves a higher BLEU score compared to other systems.

\begin{table}
\centering
\small
\begin{tabular}{l c c}
\toprule
             & DSTC2     & CamRest   \\
\midrule
$\text{uTOD}^{\text{REINFORCE}}$        & 0.02$\pm$0.01 & 0.00$\pm$0.00 \\
$\text{uTOD}^{\text{MAPO}}$           & 0.08$\pm$0.01& 0.10$\pm$0.01 \\
$\text{uTOD}^{\text{\sys}}$          & \textbf{0.17$\pm$0.02} & \textbf{0.21$\pm$0.02}\\
\midrule
$\text{aTOD}^{\text{SL+RL}}$       & 0.20$\pm$0.01  & 0.28$\pm$0.02 \\
\bottomrule
\end{tabular}
\caption{KB Entity F1 achieved by various TOD systems on OOV test set}
\label{tab:tod-oov-study}
\end{table}

\begin{table}[t]
\centering
\small
 \begin{tabular}{l cc}
\toprule
& \textbf{Info.} & \textbf{Grammar}  \\
\midrule
$\text{uTOD}^{\text{REINFORCE}}$   &  0.20 & 0.89 \\
$\text{uTOD}^{\text{MAPO}}$  & 0.36 & 1.18 \\
$\text{uTOD}^{\text{\sys}}$  & \textbf{0.64} & \textbf{1.38} \\
\midrule
$\text{aTOD}^{\text{SL+RL}}$ & 1.08 & 1.39 \\
\bottomrule
\end{tabular}
\caption{Human Evaluations on CamRest} 
\label{tab:human_eval}
\end{table}

\begin{table*}[ht]
\centering
\footnotesize
\begin{tabular}{l|p{11.5cm}}
\toprule
%\textbf{kb} & \textit{da\_vinci\_pizzeria}\\
% & \textit{r_phone|01223\_351707} \\
% & \textit{r_adddress|20\_milton\_road\_chesterton} \\
% & \textit{r_food|italian} \\
\multicolumn{2}{c}{\textbf{KB Results (Restaurant | Food | Area | Pricerange)}} \\
\multicolumn{2}{c}{\textit{nandos} | \textit{portuguese} | \textit{south} | \textit{cheap}}\\
\multicolumn{2}{c}{\textit{the\_lucky\_star} | \textit{chinese} | \textit{south} | \textit{cheap}}\\
\midrule
\midrule
\textbf{User-1} & hello , i am looking for a \textit{cheap} restaurant on the \textit{south} side of town . \\
\midrule
\textbf{Gold} & SELECT * FROM KB WHERE area = south AND pricerange = cheap \\
& there is a place called \textit{nandos} that specializes in \textit{portuguese} food on the \textit{south} side of town .\\

\midrule
\midrule
$\text{uTOD}^\text{REINFORCE}$ & SELECT * FROM table WHERE area = i \\
&  \textit{chiquito\_restaurant\_bar} is \textit{cheap} in the \textit{south} of town . would you like their phone number\\ 

\midrule
$\text{uTOD}^\text{MAPO}$  & SELECT * FROM table WHERE pricerange = cheap \\
& \textit{the\_missing\_sock} is a \textit{cheap} restaurant in the \textit{south} . would you like their phone number\\ 

\midrule
 $\text{uTOD}^\text{\sys}$ & SELECT * FROM table WHERE area = south AND pricerange = cheap \\
& \textit{nandos} are \textit{portuguese} in the \textit{south} \textit{south} would you like their phone number \\ 

\midrule
$\text{aTOD}^{\text{SL+RL}}$ & SELECT * FROM table WHERE area = south AND pricerange = cheap \\
& \textit{the\_lucky\_star} is a \textit{cheap} restaurant in the \textit{south} \textit{south} would you like their phone number \\
\bottomrule
\end{tabular}
\caption{Queries and responses generated by $\text{uTOD}^\text{\sys}$ and other TOD systems on a dialog from CamRest. For simplicity, only the fields used in the dialog are mentioned in the KB results. Entities are italicized.}
\label{tab:camrest-qualeval}
\end{table*}

\begin{table*}[ht]
\centering
\small
\begin{tabular}{c c | c c c | c c c}
\toprule
\multicolumn{2}{c|}{\textbf{Training}} & \multicolumn{3}{c|}{\textbf{DSTC2}} & \multicolumn{3}{c}{\textbf{CamRest}} \\
\cmidrule{3-8}
\textbf{Q.Position} & \textbf{Query} & \textbf{Ent. F1} & \textbf{Ent. F1} & \textbf{BLEU} & \textbf{Ent. F1} & \textbf{Ent. F1} & \textbf{BLEU} \\
\textbf{Predictor}& \textbf{Predictor} & KB & All & & KB & All & \\
\midrule
%\multicolumn{7}{c}{\textit{Query position predictor trained using gold labels}} \\
%\midrule
%Unsupr.& Unsupr. & 0.23$\pm$0.02 & 0.37$\pm$0.02 & 46.33$\pm$0.75 &          
%0.22$\pm$0.01 & 0.35$\pm$0.03 & 13.69$\pm$0.77 \\
Predicted & Predicted & 0.24$\pm$0.02 & 0.41$\pm$0.02 & 48.35$\pm$1.58 &
%0.25$\pm$0.03 & 0.36$\pm$0.02 & 13.80$\pm$1.04 \\
0.29$\pm$0.03 & 0.41$\pm$0.02 & 14.68$\pm$0.85 \\
%\midrule

Oracle & Predicted & 0.32$\pm$0.04  & 0.40$\pm$0.02 & 48.52$\pm$1.31 &        
%0.30$\pm$0.02 & 0.41$\pm$0.03 & 14.73$\pm$1.23 \\
0.32$\pm$0.02 & 0.40$\pm$0.02 & 14.17$\pm$0.70 \\

Predicted & Oracle. & 0.32$\pm$0.03  & 0.41$\pm$0.03 & 48.94$\pm$1.80 &    
0.37$\pm$0.04 & 0.44$\pm$0.03 & 14.63$\pm$0.82 \\

Oracle & Oracle& 0.38$\pm$0.03 & 0.41$\pm$0.02 & 49.79$\pm$1.80 &               
0.39$\pm$0.04 & 0.45$\pm$0.03 & 14.84$\pm$0.94 \\

\bottomrule
\end{tabular}
\caption{Performance gap for supervised TOD systems when trained in different evaluation settings.}
\label{tab:future}
\end{table*}

\noindent \textbf{Human Evaluation}: We report the human evaluation\footnote{We used two in-house (non-author) judges. One was an expert in dialog systems and the other was a novice.} results on 100 random context-response pairs for CamRest dataset in Table \ref{tab:human_eval}. $\text{uTOD}^\text{\sys}$ outperforms other $\text{uTOD}$ baselines on both informativeness and grammar. It was surprising to see $\text{uTOD}^\text{REINFORCE}$ perform poorly on grammar. Further investigation showed that often the responses generated were missing entities which made them look incomplete. For example, the response "is a restaurant in the moderate part of town . would you like their phone number" has a missing restaurant name at the start of the sentence. We measure the inter-annotator agreement using Cohen’s Kappa ($\kappa$) \cite{cohenkappa}.  The agreement was substantial for informativeness ($\kappa=0.62$) and moderate ($\kappa=0.45$) for grammar.

\noindent \textbf{Disentanglement Study}: TOD systems can learn to predict KB entities in the response either by inferring (copying) them from the query results or memorizing (generating) them. Only systems that learns to copy them from the results will generalize to entities unseen during train. To test this ability to generalize, we follow \newcite{raghu2019disentangling} and construct an OOV test set (and a corresponding KB) such that it contains entities unseen during train. Table \ref{tab:tod-oov-study} shows KB entity F1  of various TOD systems on this OOV test set. The low numbers for $\text{uTOD}^\text{REINFORCE}$ confirm that it memorizes KB entities. 
$\text{uTOD}^\text{\sys}$ achieves best scores, and is 3-7 points short of  supervised TOD.

\noindent \textbf{Qualitative Example}
We qualitatively compare the performance of various TOD systems using the example shown in Table \ref{tab:camrest-qualeval}. The example demonstrates the ability of $\text{uTOD}^\text{\sys}$ to generate the correct query and use the KB results to generate an appropriate response. $\text{uTOD}^\text{MAPO}$ generates a partial query which has a larger set of results – this confuses the response predictor, and hence an incorrect restaurant is copied in the response. $\text{uTOD}^\text{REINFORCE}$ fails to generate a valid query and so the response predictor generates a random restaurant memorized during train. The response generated by all approaches contain the correct context entities (\textit{cheap} and \textit{south}), but only responses of $\text{uTOD}^\text{\sys}$ and $\text{uTOD}^\text{SL}$ contain the appropriate KB entity requested by the user.

\section{Discussion \& Research Challenges}

We now discuss novel research questions identified in this research.

\emph{Can we train an end-to-end differentiable uTOD system?} Our proposed approach for uTOD uses a pipeline of three components trained separately, which can lead to cascading errors. We believe that a single end-to-end neural architecture will likely obtain superior performance. But, this is a technical challenge, since it will require one model to make two discrete decisions -- when to query and what to query, complicating the RL problem. 

\begin{comment}
\emph{Can we combine RL with supervised loss for training a better aTOD system?} We make the surprising observation in Table \ref{tab:performace-study} that on CamRest, supervision is actually hurting overall performance due to overfitting. We believe that it might be possible to carefully combine weak supervision using RL as a secondary source of knowledge on top of cross entropy loss for better training. This is analogous to the benefits of additional prior knowledge (e.g., as constraints) on top of supervision for small datasets \cite{nandwani-nips19}. However, our preliminary experiments using \emph{total loss = Cross-Entropy - }$\lambda.$\emph{reward} did not achieve good results. To the best of our knowledge, there is no work that uses RL as a regularizer for supervised training.
\end{comment}

\emph{Can we bridge the gap between aTOD performance with and without oracle queries at test time?} Our work exposes a critical weakness of fully supervised TOD systems. When evaluated in a setting where the TOD system only has access to its own predicted query and that query's results, the overall performance drops drastically. Table \ref{tab:future} shows the performance of our aTOD system, where at test time, both for query position and the query, the system prediction is used (Prediction) or the gold value is used (Oracle). We observe that using oracle values gets 14 pt KB Entity F1 gains over predicted values. Since our setting is very realistic to assess the usability of current dialog systems, this result highlights the research challenge of improving supervised TOD systems in our setting. We also emphasize the importance of KB Entity F1 as an important metric --  it better assesses the value offered to the end user, given that these datasets contain mostly informational tasks.

\emph{Can we design a uTOD system to handle dialogs that require more than one query per dialog?} Our current work makes the assumption of a single query per dialog. This is because most supervised TOD datasets also make only one query per dialog. Given that our task is harder than those, it was important to define it such that existing ML machinery of TOD can feasibly learn our task. At the same time, extending the task definition and creating datasets for the multiple queries per dialog case is straightforward. An important future research challenge will be to design an end-to-end dialog system that can handle a variable number of KB queries in a single dialog, without explicit query (or query position) annotation. We believe that this is best studied only after substantial progress on our specific task definition.

\section{Conclusion}
We define the novel problem of learning TOD systems without explicit KB query annotation and the associated subtask of unsupervised prediction of KB queries. We also propose first baseline solutions for these tasks. Our best query prediction baseline extends existing RL approach MAPO to include multiple query buffers at training time.  We also present a pipeline architecture that trains different components in a curriculum to obtain the final TOD system. Our detailed evaluation shows that our approaches achieve much better performance than simpler baselines, though there is some gap when compared to supervised approaches. We study the results further to identify research challenges for future research.
%, which include designing a single architecture for our task, and improving the standard supervised TOD systems in our novel evaluation setting.
We will release all resources for use by the research community.

%We propose a reinforcement learning(RL) baseline for unsupervised query predcition which makes simple modifications to MAPO to counter correlated attributes and unequal rewards.  We propose a pipelined approach which independently predicts the KB query, and uses the query results to predcit responses in subsequent dialog. Finally, we elaborate the issue with existing TOD evaluation setup and describe future directions for learning TOD system without explicit KB query annotations. We will release all our resources for further research -- training data, evaluation code, and code for TOD systems.

\section*{Acknowledgments}
We thank Gaurav Pandey, Danish Contractor, Dhiraj Madan, Sachindra Joshi and the anonymous reviewers for their comments on an earlier version of this paper. This work is supported by IBM AI Horizons Network grant, an IBM SUR award, grants by Google, Bloomberg and 1MG, a Visvesvaraya faculty award by Govt. of India, and the Jai Gupta chair fellowship by IIT Delhi. We thank the IIT Delhi HPC facility for computational resources.

\bibliography{tacl2018}
\bibliographystyle{acl_natbib}

\end{document}